\documentclass[10pt,twocolumn,letterpaper]{article}

\usepackage[pagenumbers]{cvpr} 

\usepackage{graphicx}
\usepackage{amsmath}
\usepackage{amssymb}
\usepackage{booktabs}
\usepackage{xcolor}
\usepackage{paralist}
\usepackage{caption}
\usepackage{lipsum}

\usepackage{makecell}

\newcommand{\thename}{Get3DHuman}
\newcommand{\modelname}{Get3DHuman}

\usepackage[pagebackref,breaklinks,colorlinks]{hyperref}

\begin{document}

\title{\textit{Get3DHuman}: Lifting StyleGAN-Human into \\ a 3D Generative Model using Pixel-aligned Reconstruction Priors}

\author{Zhangyang Xiong$^{1,2}$ \quad Di Kang$^{3}$ \quad Derong Jin$^{2}$  \quad Weikai Chen$^{4}$ \quad Linchao Bao$^{3}$ \\ \quad Shuguang Cui$^{2,1}$\quad Xiaoguang Han$^{2,1*}$ \\
{\normalsize $^1$FNii, CUHKSZ}\quad
{\normalsize $^2$SSE, CUHKSZ}\quad
{\normalsize $^3$Tencent AI Lab}\quad
{\normalsize $^4$Tencent America}
}


\twocolumn[{
\maketitle
\begin{center}
\vspace{-6mm}
    \captionsetup{type=figure}
    \includegraphics[width=1.0\textwidth]{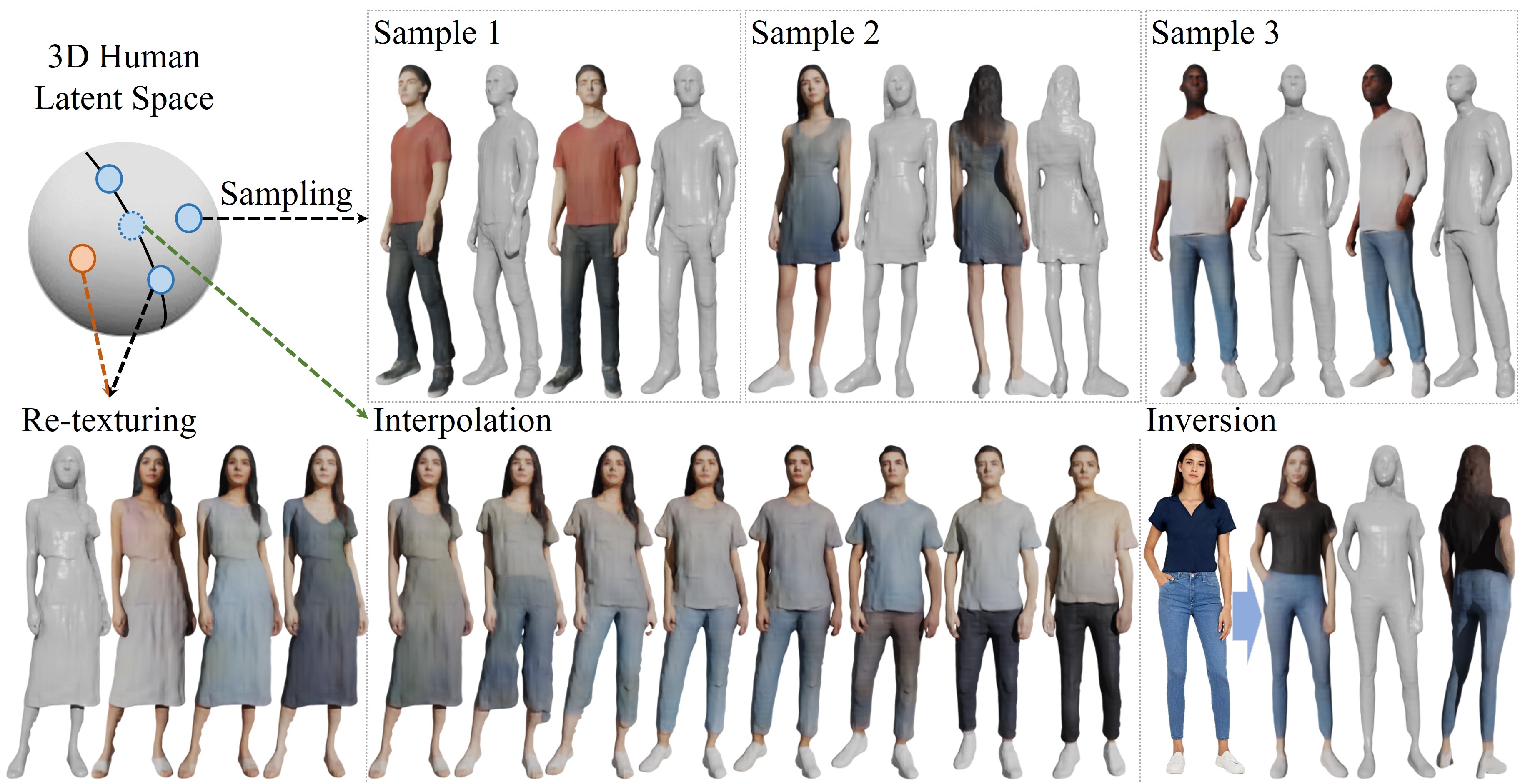}
    \caption{\textbf{Generations from {\thename}.}
    We export generated shapes and visualize them in Blender.  
    Besides generating 3D textured human models from random codes, our method also supports re-texturing a given shape (bottom left), shape and texture interpolation (bottom middle), and inversion from a given reference image (bottom right). More results show from Fig.~\ref{fig:mul_view} to ~\ref{fig:sup_inter_1}.}
\end{center}
}]
    



\begin{abstract}
Fast generation of high-quality 3D digital humans is important to a vast number of applications ranging from entertainment to professional concerns. 
Recent advances in differentiable rendering have enabled the training of 3D generative models without requiring 3D ground truths.
However, the quality of the generated 3D humans still has much room to improve in terms of both fidelity and diversity.
In this paper, we present \emph{\modelname{}}, a novel 3D human framework that can significantly boost the realism and diversity of the generated outcomes by only using a limited budget of 3D ground-truth data.
Our key observation is that the 3D generator can profit from human-related priors learned through 2D human generators and 3D reconstructors.
Specifically, we bridge the latent space of \modelname{} with that of StyleGAN-Human~\cite{fu2022styleganhuman} via a specially-designed prior network, where the input latent code is mapped to the shape and texture feature volumes spanned by the pixel-aligned 3D reconstructor~\cite{saito2020pifuhd}.
The outcomes of the prior network are then leveraged as the supervisory signals for the main generator network.
To ensure effective training, we further propose three tailored losses applied to the generated feature volumes and the intermediate feature maps. 
Extensive experiments demonstrate that \modelname{} greatly outperforms the other state-of-the-art approaches and can support a wide range of applications including shape interpolation, shape re-texturing, and single-view reconstruction through latent inversion.     

\end{abstract}

\section{Introduction} \label{sec:intro}

Generating diverse and high-quality virtual humans plays an important role in numerous applications, including VR/AR, visual effects, game production, etc.
The advances in generative models have brought impressive advances to the state-of-the-art of generating 2D virtual avatars, such as images or videos.
However, synthesizing 3D humans with high fidelity and large variations remains much under-explored due to the scarcity of 3D human data.

The conventional solution~\cite{con_0, con_1, con_4, conv_3} for acquiring a 3D avatar from a real person is typically a time-consuming and cumbersome process, requiring a specialized capture system, substantial manual efforts, and extensive computation.
To circumvent the requirement of collecting a large corpus of 3D ground-truth data, recent works utilize the differentiable  rendering technique to train a 3D generative model in a 3D-unsupervised manner~\cite{Chan2022EG3D, gao2022get3d}.
Specifically, instead of using direct 3D supervision, adversarial losses are applied to the images rendered from the synthesized 3D content.
However, due to the lack of dense multi-view images for each model, these methods can only encourage geometry-to-image consistency in the selected views while failing to produce plausible reconstruction in the unseen regions.
In addition, the differentiable rendering process is computationally heavy, making the network training highly inefficient.

To resolve the above issues, in this work, we present \emph{\modelname{}}, a novel 3D generator that can faithfully synthesize high-fidelity clothed 3D humans with a diversity of shapes and textures.
Our key observation is that 3D human generators can benefit from the inductive bias from a 2D human synthesizer and the prior knowledge learned through relevant 3D modeling tasks.
In particular, to bypass the limited availability of 3D ground truths, we propose to leverage the generative power of 2D human synthesizers which have shown more promising and stable quality than their counterpart 3D generators.
We further lift the rich prior from the 2D generator, i.e. StyleGAN-Human~\cite{fu2022styleganhuman}, to 3D by using the pixel-aligned reconstruction priors, i.e. the pre-trained PIFu network~\cite{saito2019pifu}, through single-view human reconstruction.
Thanks to the strong generalization ability of pixel-aligned implicit reconstructor, by feeding it with a myriad of photorealistic human images generated by StyleGAN-Human, we are able to obtain a vast number of 3D human models with highly diversified body shapes, apparel, poses, and textures.
To further ensure the high quality of the generated shapes, we filter out inferior results via manual inspections.

\begin{figure*}[th]
\centering
\includegraphics[width=0.99\linewidth]{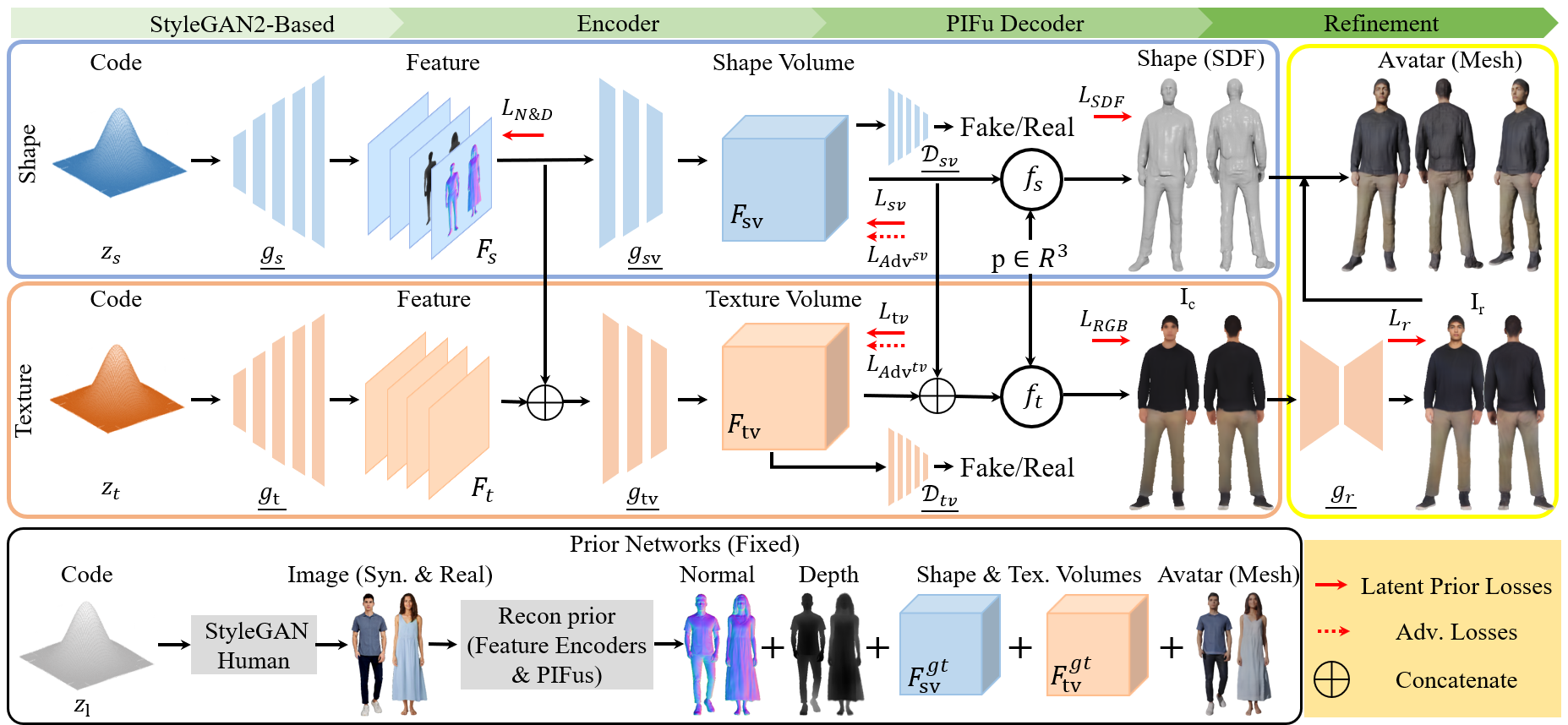}
\vspace{-2mm}
\caption{\textbf{Overview of our framework.} 
Our {\thename} consists of a shape generator (blue box) and a texture generator (orange box) with a refinement module (yellow box) that enables nonexistent 3D human creation. 
Shape generator responds for generating a high-quality full-body geometry from a shape code and sends shape features to the texture generator. 
Texture generator predicts RGB colors of all points in the 3D space from a texture code and intermediate shape features.
Trainable modules are underlined, including $g_s$, $g_{sv}$, $\mathcal{D}_{sv}$, $g_t$, $g_{tv}$, $\mathcal{D}_{tv}$, and $g_r$. These seven modules are all trained from scratch. The prior network (black box) only produces supervisory signals for the training of form.}
\label{fig:pipeline}
 \vspace{-2mm}
\end{figure*}

We further materialize the above idea via a novel prior induction mechanism.
Specifically, we first train a prior network to encode the 2D generator prior and the 3D reconstruction prior into three supervisory signals.
That is, given a random latent code, the prior network would generate normal maps, depth maps, and shape and texture feature volumes of the 3D human corresponding to the input code (see Fig.~\ref{fig:pipeline}).
The input code is sampled from the latent space of StyleGAN-Human while the shape and texture feature volumes are consistent with the PIFu latent, and, hence, can be converted to the predicted human shape via the pre-trained PIFu decoder.
We then supervise the training of the proposed 3D generators via three specially-tailored losses.
First, a latent prior loss is introduced to provide direct supervision of the generated feature volumes for the shape and texture generation branches. 
Second, an adversarial loss is applied to the 3D feature volumes instead of the output signed distance field (SDF).
This helps reduce the training complexity while ensuring the realism of the generated 3D humans.
Lastly, normal maps and depth maps are used for supervising the generation of intermediate feature maps.
Specifically, instead of directly transforming the input code into a 3D feature volume, we first map the code to 2D feature maps and then lift them into 3D feature volume. 
This additional intermediate supervision helps cast finer-grained geometry details as shown in our experiments (see Fig.~\ref{fig:normal_add}).
We further utilize a refinement module to improve the quality of our textured mesh as the texture prior  is not always satisfactory.

Our method can support a wide range of applications, including shape generation, interpolation, shape re-texturing, and latent inversion from a single image.
We evaluate \modelname{} via extensive experiments and demonstrate that it strongly outperforms the state-of-the-art methods, both qualitatively and quantitatively.

To summarize, our main contributions include:
\begin{compactitem}
\item We propose a novel 3D human generation framework that explicitly incorporates priors from top-tier 2D human generators and 3D reconstruction schemes to achieve high-quality and diverse 3D clothed human generation.  
\item We present specially-tailored prior induction losses for effective and efficient prior-based supervision. 
\item We set the new state of the art in the task of shape generation while supporting many applications including shape interpolation, re-texturing, and latent inversion. 
\end{compactitem}

\section{Related Work} \label{sec:related}


Constructing generative models for images, videos or 3D models is a fundamental problem in the field of computer graphics and computer vision. Here, we only review and summarize the work related to 2D or 3D human generation that is relevant to our work.

\noindent\textbf{2D Human Generation.}
In recent years, generative adversarial networks (GANs)~\cite{gan} has been successfully applied to the task of image synthesis.
Many methods have built photorealism generative models of 2D human faces~\cite{StyleGAN1, StyleGAN2, StyleGAN3}. Based on the success of these efforts, faces can also be efficiently edited ~\cite{roich2021pivotal, NeuralFace2017, jiang2021talk}.
On another line, some researchers attempt to generate 2D images of dressed humans from sketch, pose, or text conditions \cite{pose2image2017ma, g2dpeople2017Christoph, albahar2021pose, sarkar2021humangan,text2humanjiang2022}.
The recent works of \cite{fu2022styleganhuman, Fruehstueck2022InsetGAN} have built large-scale datasets and resorted to StyleGAN to achieve impressive 2D human generation results. 
The success in 2D image generation and editing has inspired research in 3D generation.

\noindent\textbf{3D Human Modeling.}
As a common representation, parametric human model, like~\cite{anguelov2005scape, loper2015smpl, pons2015dyna, bogo2016keep, hmrKanazawa17,joo2018total, omran2018neural, pavlakos2019expressive}, controls a template through a series of low-dimensional parameters to obtain a 3D naked human body, with which the modeling can be performed by regression approaches. These works are further extended to model dressed human 
\cite{clothsmpl12018Alldieck, Dress2020Ma}, by introducing vertices' movements on the template meshes.
Some other representations are also proposed for dealing with more complex clothed bodies, like point clouds~\cite{scale2021ma, POP_2021, Zakharkin_2021_ICCV}, radiance fields~\cite{liu2021neuralac, 2021narf, peng2021animatable, su2021anerf} and implicit fields~\cite{chen2019learning,mescheder2019occupancy,park2019deepsdf,xu2019disn}.
In particular, the implicit field representation has become a powerful tool for modeling 3D (clothed) human shapes, as it can capture arbitrary resolutions and fine-grained details.

PIFu~\cite{saito2019pifu} first proposes a pixel-aligned implicit function to extract local features from images to complete human body digitization.
They are further expanded to PIFuHD~\cite{saito2020pifuhd}, which uses predicted normal maps and a multi-level architecture to generate higher resolution and richer details consistent with the input image.

\noindent\textbf{3D Human Generation.}
Early approaches aimed to extend GANs to voxel~\cite{wu2016learning, nguyen2020blockgan}, point cloud~\cite{pc1_2018, zhou20213d_pc2, mo2019structurenet_pc3}, implicit~\cite{mescheder2019occupancy, chen2019learning}.
However, these works focus mainly on generating geometry, not appearance. 

EG3D~\cite{Chan2022EG3D} proposes to generate multi-view images of a 3D face via introducing an efficient tri-plane representation and a neural rendering design with the help of a pretrained StyleGAN.

Following EG3D, there are also some methods aiming to full-body human generation \cite{zhang2022avatargen, EVA3D}. Since their adversarial losses are only applied to the rendered images and there is no explicit supervision on the geometry, making the produced shape is far from satisfactory. 

Similar to our work, GET3D~\cite{gao2022get3d} also targets to generate textured 3D meshes. Specifically, it generates an SDF field and a texture field by two latent codes and performs simultaneous optimization of the two fields by adversarial loss on 2D rendered images. 
In our work, we focus on 3D textured human models and aim to generate 3D textured meshes with diverse human poses and appearances.  

Recently, a concurrent work HumanGen~\cite{HumanGen} is posted on the Internet. Although it also involves the prior of StyleGAN-Human and PIFu for generative model construction, we have a key difference: HumanGen only builds a generative model to produce a texture field from a code where the geometry is directly borrowing from the prior, while ours aims to use two codes to generate both shape field and texture field. In this fashion, different from HumanGen, we truly built a latent space for 3D textured humans.


\section{Method} 
\label{sec:method}
%

%
Fig.~\ref{fig:pipeline} shows the overall network of the proposed 3D generator network.
{\thename} contains parallel shape and texture branches, which is facilitated by a prior network branch during training.
%
Take the shape branch as an example, given a shape latent code, {\thename} generates a shape feature $F_s$, a shape feature volume $F_{sv}$, and produces a high-quality human shape represented as an SDF (Sec.~\ref{sec:Generator_Architecture}).
Sec.~\ref{sec:prior_Supervision} details our prior-facilitated learning of this 3D generator.
%
Sec.~\ref{sec:Texture_G} describes the texture network and its training.
%
Finally, training details including data preparation, and training strategy are described in Sec.~\ref{sec:Training}.

\subsection{The Proposed 3D Generator} 
\label{sec:Generator_Architecture}

%
As shown in Fig.~\ref{fig:pipeline} top, it contains three parts, including a StyleGAN-like Mapping \& Synthesis stage ($g_{s}$), a feature encoder stage ($g_{sv}$), and a PIFu-style geometry decoder ($f_{s}$).
%
Given a shape latent code ${z}_s$ drawn from a Gaussian distribution, 
we first use a StyleGAN2 network~\cite{StyleGAN2} $g_s$ to generate a shape feature $F_s$,
which is fed into a hourglass-style~\cite{newell2016hourglass} fully convolutional feature encoder $g_{sv}$ to extract a shape feature volume $F_{sv}$.
\ie:
\begin{gather}
\label{eq:shape_gen1}
\setlength{\abovedisplayskip}{3pt}  
\setlength{\belowdisplayskip}{3pt}  
    F_{s} = g_{s}(z_{s}), \\
    F_{sv} = g_{sv}(F_{s}),
\end{gather}
where $z_{s}$ is the random shape latent code, $F_{s}$ and $F_{sv}$ are the intermediate pixel-aligned high-dimensional \textbf{s}hape feature image and \textbf{s}hape feature volume, respectively. 

With the shape feature volume, a PIFu~\cite{saito2019pifu} shape decoder $f_s$ is used to evaluate the signed distance $s\in\mathbb{R}$ given a query point $p\in\mathbb{R}^3$ and its corresponding feature.
\ie:
\begin{align}
\label{eq:shape_gen3}
\setlength{\abovedisplayskip}{3pt}  
\setlength{\belowdisplayskip}{3pt}  
    f_{s}(F_{sv}(x), d(p)) = s : s\in\mathbb{R},
\end{align}
where $p \in \mathbb{R}^3$ is the 3D query point, 
$F_{sv}(x)$ is the image feature at $p$'s projected location $x=\pi(p)$ on the image plane,
$d(p)$ is the depth value of $p$ in the camera frame, 
and $f_{s}$ is the PIFu shape decoder (MLPs) that is pretrained like \cite{saito2020pifuhd} and fixed.

\subsection{Prior-Facilitated Learning} 
\label{sec:prior_Supervision}
Besides adversarial loss, we propose to fully utilize the prior information from well-trained networks to facilitate the training of the 3D GAN 
due to limited training data and the complexity of this task.
Specifically, we first utilize image-latent pairs from StyleGAN-Human to learn the latent space better since it already has a reasonable structure after training.
Then, we extract intermediate features using the PriorNet branch supervision to guide the training, which can also be seen as a kind of deep supervision.
In the following, we first describe how to extract helpful prior information and then describe the training losses.

\vspace{1mm}
\noindent\textbf{Prior extraction.}
The prior network is the concatenation of a StyleGAN-Human~\cite{StyleGAN2} generator and  a PIFuHD-like~\cite{saito2019pifu,saito2020pifuhd} 3D reconstructor (see Fig.~\ref{fig:pipeline}).
So, given a latent code,
a full body human image is first synthesized by the StyleGAN-Human generator.
Then the following 3D reconstructor takes as input the synthesized image and extracts a normal map $N$, a depth map $D$, a shape feature volume $F_{sv}$, and a texture feature volume $F_{tv}$, which are used as supervisory signals since they contain helpful human prior information.

\vspace{1mm}
\noindent\textbf{Shape losses.}
Given a latent code, the shape generator produces intermediate shape feature $F_{s}$ and shape feature volume $F_{sv}$, and the final SDF $s$ (see Fig.~\ref{fig:pipeline}).
Our training losses include a latent adversarial loss $\mathcal{L}_{Adv^{sv}}$ applied on the shape feature volume $F_{sv}$, 
a SDF loss $\mathcal{L}_{SDF}$,
a latent prior loss $\mathcal{L}_{sv}$ applied on the shape feature volume, a depth loss $\mathcal{L}_{D}$ and a normal loss $\mathcal{L}_{N}$ applied on the first four channels of the intermediate shape feature $F_{s}$. 
The loss terms are detailed in the following.

\textit{SDF loss $\mathcal{L}_{SDF}$.}
We sample $M$ points per iteration, including near surface points (obtained from depth map) and random points, and apply L1 loss on them. 
The SDF value is predicted by a pretrained and fixed PIFu MLP decoder~\cite{saito2019pifu} 
$s = f_{s}(F_{sv}(x), d(p_{i}))$, where $x=\pi(p)$ is the 2D projection of query point $p$ and $d(p)$ is the depth value of $p$ in the camera coordinate space.
\begin{equation}
\setlength{\abovedisplayskip}{3pt}  
\setlength{\belowdisplayskip}{3pt}  
\mathcal{L}_{SDF} = \frac{1}{M}\sum_{i}^{M}||\hat{s} - s||_1
\end{equation}

\textit{Latent prior loss $\mathcal{L}_{sv}$.}
To take full use of the prior knowledge, we also apply constraints on the intermediate features as follows. 
Another potential benefit is the sampled 3D points in the previous SDF loss are too sparse while this feature field loss can provide supervision on the whole feature maps.
%
\begin{equation}
\mathcal{L}_{sv} = ||F_{sv} - F_{sv}^{gt}||_1
\end{equation}

\textit{Geometry loss.} 
We force several channels of the intermediate feature maps to predict helpful 2.5D information (i.e. normal) since these feature maps are pixel-aligned (i.e. the spatial information is retained).
This operation is similar to deep supervision~\cite{lee2015deeply} and helps the learning processing.
\begin{equation}
\begin{split}
\mathcal{L}_{N\&D} = & 
\lambda_{N}||\hat{F}_{s(c:1,2,3)} - N||_1  + 
\\& \lambda_{D}||\hat{F}_{s(c:4)} - D||_1  
\end{split}
\end{equation}
where $\hat{F}_{s(c:1,2,3)}$ and $\hat{F}_{s(c:4)}$ are the first three and fourth channels of the predicted shape feature $F_s$, $D$ and $N$ is the pseudo groundtruth normal map and depth map.

\textit{Latent adversarial loss $\mathcal{L}_{Adv^{sv}}$.}
$\mathcal{D}_{sv}$ is a discriminator taking the shape volume as input.
We use the non-saturating GAN loss proposed in StyleGAN2~\cite{StyleGAN2} and R1 regularization~\cite{gan, l1reg}. 
\begin{equation} 
\mathcal{L}_{Adv^{sv}} = \mathcal{L}_{GAN^{sv}} + \lambda_{Reg} \mathcal{L}_{Reg^{sv}}
\end{equation}

In summary, the total shape loss $\mathcal{L}_{Shape}$ is
\begin{equation}
\label{eq:shape_loss}
\begin{split}
\mathcal{L}_{Shape} = & 
\lambda_{SDF} \mathcal{L}_{SDF} +
\lambda_{sv} \mathcal{L}_{sv} + 
\lambda_{N} \mathcal{L}_{N} + 
\\ & 
\lambda_{D} \mathcal{L}_{D} + 
\lambda_{Adv^{sv}} \mathcal{L}_{Adv^{sv}} 
\end{split}
\end{equation}
where $\lambda_{(\cdot)}$s are the corresponding loss weights.
After training, the shape generator is fixed during the later texture branch learning.

\subsection{Texture Generator \& Training Losses} \label{sec:Texture_G}

The texture branch is almost a mirror of the shape branch except for several small differences.

\noindent\textbf{Texture generator.}
The texture branch consists of a StyleGAN-style generator $g_t$, a feature encoder $g_{tv}$, a PIFu-style texture decoder $f_t$, and a texture volume discriminator $\mathcal{D}_{tv}$.
The texture branch is similar to the shape branch except it is additionally conditioned on intermediate features from the shape branch (see Fig.~\ref{fig:pipeline} and Eqs.~\ref{eq:shape_gen1}-\ref{eq:shape_gen3}).
\ie:
\begin{gather}
\setlength{\abovedisplayskip}{3pt}  
\setlength{\belowdisplayskip}{3pt}  
    F_{t} = g_{t}(z_{t}),  \\
    F_{tv} = g_{tv}(F_{t} \oplus F_{s}), \\
    f_{t}((F_{tv}(x) \oplus F_{sv}(x)), d(p)) = c \in [0, 1]^3,
\end{gather}
where $z_{t}$ is texture latent code, 
$\oplus$ represents channel-wise concatenation to introduce intermediate features (i.e. $F_{s}$, $F_{sv}(x)$) from the shape branch.
Similarly, $f_{t}$ is the PIFu texture decoder (MLPs) that is pretrained like \cite{saito2019pifu} and fixed.
The texture discriminator is also similar to the shape discriminator but with 512 input channels.

\vspace{1mm}
\noindent\textbf{Texture losses.}
Similar to the shape branch, given a texture latent code, the texture branch first generates a texture feature ($F_{t}$) and a texture feature volume ($F_{tv}$).
Different from the shape branch, it also takes as input intermediate shape features (e.g. $F_s$, $F_{sv}$) that are generated using the same latent code $z$ (i.e. $z_s$ = $z_t$).
At the same time, the texture branch use the same latent code to predict a texture feature ($F_{t}$). 
Similarly, the individual texture losses are  as follows:
\begin{gather}
\setlength{\abovedisplayskip}{3pt}  
\setlength{\belowdisplayskip}{3pt}  
\mathcal{L}_{RGB} =  \frac{1}{M}\sum_{i}^{M}||\hat{c} - c||_1,  \\
\mathcal{L}_{tv} = ||F_{tv} - F_{tv}^{gt}||_1, \\
\mathcal{L}_{Adv^{tv}} =  \mathcal{L}_{GAN^{tv}} + \lambda_{Reg} \mathcal{L}_{Reg^{tv}},
\label{eq:texture_losses}
\end{gather}
%
The overall texture loss $\mathcal{L}_{Texture}$ is:
\begin{equation}
\mathcal{L}_{Texture} \;=\;  \lambda_{RGB} \mathcal{L}_{RGB} \;+
\lambda_{tv} \mathcal{L}_{tv} \;+ 
\lambda_{Adv^{tv}} \mathcal{L}_{Adv^{tv}}
\label{eq:total_texture_loss}
\end{equation}
where $\lambda_{(\cdot)}$s are the corresponding loss weights.
Note that we fix the shape branch and train the texture branch.

\subsection{Refinement Module} \label{sec:refine}
A textured mesh can be obtained from the two implicit fields similar to PIFu~\cite{saito2019pifu}.
%
However, due to limited training data, the reconstruction prior learned in the texture field is not always satisfactory.
For example, the rendered images could be blurry or sometimes erroneous, which means the corresponding textured meshes extracted from the implicit fields could also be problematic.

To obtain high-quality mesh colors, we propose an extra refinement module, which consists of an image refinement step and a (mesh) texture update step.

%
%

\noindent\textbf{Image refinement.}
The image refinement is realized as an image-to-image (I2I) translation task using a UNet-style~\cite{jafarian2021learning} network $g_{r}$.
\begin{equation}
\label{eq:refine}
\setlength{\abovedisplayskip}{3pt}  
\setlength{\belowdisplayskip}{3pt}  
I_{r} = g_{r}(I_{c})
\end{equation}
where $g_{r}$ is the network,
$I_{c}$ is the rendered image from shape and texture volumes, 
and $I_{r}$ is the refined image for later texture extraction.
We impose L1 loss and perceptual loss~\cite{perceptualloss} between the refined result and the ground truth image $I^{gt}$ during training. 
The loss $\mathcal{L}_{r}$ is defined as:
%
\begin{equation}
\begin{split}
\mathcal{L}_{\text {r}}= &
\lambda_{r} ||I_{r} - I^{gt}||_1 + 
\\& \lambda_{P} \sum_l\left\|\Phi^l(I_{r})-\Phi^l(I^{gt})\right\|_2^2,
\end{split}
\label{eq:I2I_loss}
\end{equation}
where $\Phi^l$ denotes the multi-level feature extraction operation using a pretrained VGG network, 
$\lambda_{(\cdot)}$s are the corresponding loss weights.

After image refinement, we can get an person image in the same pose but with sharper and more correct face/cloth details (see Fig.~\ref{fig:refine}).
%

\noindent\textbf{Vertex-color refinement.}
With the improved multi-view images, we can paint the extracted mesh $\mathcal{M}$ with new colors accordingly.
%
Specifically, we utilize surface tracking~\cite{liu2020dist} to render multi-view depth maps paired with those images.
%
With paired depth maps and their corresponding refined images, we can obtain a high-quality colored point cloud $\mathcal{P}$ via 2D-to-3D projection.
Finally, for every mesh vertex $v$, we paint it with the color of its nearest neighbor point in the colored point cloud $\mathcal{P}$.

%
%



\subsection{Data Preparation \& Training} \label{sec:Training}

\noindent\textbf{Collecting \& filtering data.}
Three different types of data are used in this work, including 396 high-quality 3D human models, real-world Internet images, and StyleGAN-Human~\cite{fu2022styleganhuman} synthesized images.
More concretely, the 3D human models are purchased from RenderPeople~\cite{renderpeople}, which include both 3D mesh models and high-resolution texture maps that could be rendered into photorealistic images.
%
%
The real images are crawled from the Internet and 14,097 are left after a manual selection to exclude extreme poses and oddly shaped costumes that can not be handled by reconstruction prior.
%
The synthesized images are generated using StyleGAN-Human~\cite{fu2022styleganhuman} and 69,099 are left after a manual selection to exclude images with obvious artifacts (e.g. distorted body parts). 
%
%

%
\noindent\textbf{Extracting pseudo-GT.}
We extract pseudo-GT for a given image using two prior networks, a human image synthesis network~\cite{zheng2022sdfstylegan} and a single-view reconstruction network~\cite{saito2019pifu}, which are trained using our RenderPeople data.
For a real image, the corresponding pseudo-GT includes a real image, a depth map, a normal map, the shape volume $\mathcal{F}_{sv}$, and the texture volume $\mathcal{F}_{tv}$.
For a synthesized image, we also stored its corresponding latent code.
A pretrained PIFu decoder can easily extract 3D or texture information from specific feature volumes.
After the aforementioned pre-processing, we randomly choose 500 (real) and 1500 (synthesis) pseudo-GT for evaluation and use the remaining 81,196 samples during training. 



\noindent\textbf{Network training.}
We train the shape branch, the texture branch, and the refinement network in three separate stages.

In the $1st$ stage, we only train the shape branch (with the  PIFu shape decoder $f_s$ fixed).
Specifically, we train the shape branch using the shape reconstruction losses in Eq.~\ref{eq:shape_loss}.
The $\lambda_{(\cdot)}$s are set to $\{20, 40, 20, 20, 1\}$, empirically. 
We found the weight of the adversarial loss has an obvious influence on the synthesized 3D models.
For example, using too large adversarial loss usually produces unrealistic high-frequency noise, and using too small adversarial loss usually results in overly-smoothed 3D models, showing similar trends to Fig.~\ref{fig:gan_prior}.

In the $2nd$ stage, we train the texture branch with the shape branch and the PIFu texture decoder $f_t$ fixed.
The prior losses $\mathcal{L}_{RGB}$, $\mathcal{L}_{tv}$ and the adversarial loss $\mathcal{L}_{adv^{tv}}$ in Eq.~\ref{eq:total_texture_loss} are applied simultaneously.
The $\lambda_{(\cdot)}$s are set to $\{20, 40, 1\}$, empirically.
Note that examples generated with both paired shape-texture latent codes (i.e. $z_s = z_t$) and unpaired latent codes (i.e. $z_s \neq z_t$) will be used for  adversarial training so that we can obtain different and reasonable textures for the same shape latent code, which enables the re-texturing application.

In the $3rd$ stage, we only train the I2I refinement network $g_{r}$ with all the other parts fixed.
The L1 loss $\mathcal{L}_{r}$ and the perceptual loss $\mathcal{L}_{P}$ in Eq.~\ref{eq:I2I_loss} are applied simultaneously.
The $\lambda_{(\cdot)}$s are set to $\{1, 1\}$.
Note that $360^{\circ}$ images rendered from RenderPeople data are additionally used for training in this stage.

\begin{table}[tb]
\caption{Quantitative comparisons with SOTA methods.}
\small
\label{tab:sota}
{\def\arraystretch{1} \tabcolsep=0.38em 
\begin{tabular}{r|cccccl}
\toprule
 & COV$\uparrow$ (\%) & MMD$\downarrow$ & FPD$\downarrow$ & FID$\downarrow$ & $\text{FID}_\text{3D}$ $\downarrow$ \\
\midrule
EG3D &  15.33 & 2.54 & 2.21 & 76.55 &  221.73\\
SDF-StyleGAN &  23.35 & 1.12 & 1.02  & - & - \\
GET3D &  35.93 & 0.77 & 0.87 & 61.69 & 88.15\\
\midrule
Ours &  \textbf{39.22} & \textbf{0.58} & \textbf{0.85} &
\textbf{54.39} & \textbf{69.70} \\
\bottomrule
\end{tabular}
}
\end{table}

\begin{figure}[tb]
\centering
\includegraphics[width=0.99\linewidth]{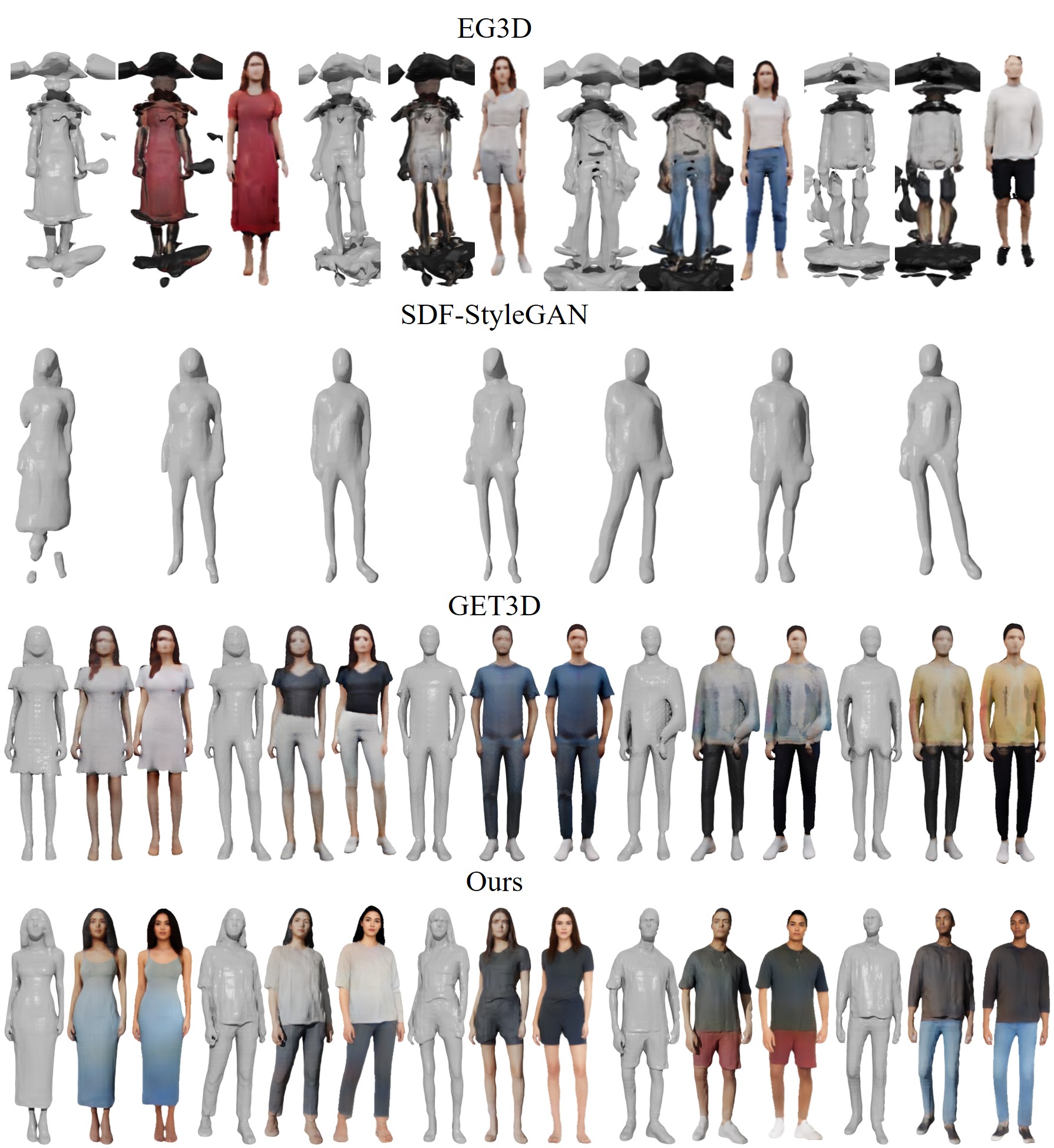}
\vspace{-4mm}
\caption{
Visual comparisons with state-of-the-art 3D human generators.
SDF-StyleGAN can only generate geometry. 
For the other methods, we visualize the geometry and appearance rendered with Blender, and the images (rightmost of each set of results) generated directly by the network. 
EG3D use volume rendering to generate images, while our method and GET3D query the RGB values on the surfaces.
Compared to the others, our results contain sharper and more plausible details in both geometry and texture, achieving the best scores on all the metrics in Tab.~\ref{tab:sota}.
\vspace{-4mm}
}
\label{fig:compare_sota}
\end{figure}

\section{Experiments} \label{sec:Experiments}
In Sec.~\ref{sec:metrics}, we first introduce the quantitative evaluation used in our experiments.
In Sec.~\ref{sec:sotas}, we compare the proposed method with other state-of-the-art methods, showing that our method generates more diverse and higher-quality textured meshes.
We conduct ablation studies of our method to verify the effectiveness of the adversarial loss, the reconstruction prior, and the refinement module in Sec.~\ref{sec:Ablation}.
Finally, we demonstrate three downstream applications of our method in Sec.~\ref{sec:applications}, including re-texture the generated meshes, interpolation between two latent codes, and inversion from the real-world image.

\noindent\textbf{Geometry evaluation.} \label{sec:metrics}
Similar to any 3D GAN, we adopt Fréchet point cloud distance (FPD)~\cite{FDP} to evaluate the diversity and quality of the generated shapes.
We use Chamfer Distance based ($d_{CD}$) Coverage (COV) and Minimum Matching (MMD) following~\cite{pc1_2018,gao2022get3d} to evaluate the similarity between a set of generated meshes and the reference set of pseudo-GT meshes. See our Supp. for detailed settings.


\noindent\textbf{Texture evaluation.}
To evaluate the quality of the generated textures, we used the common Fréchet Inception Distance (FID) metric on 2D images directly generated from the model and FID$_{3D}$ on 2D images rendered using the textured 3D mesh model. 
we randomly choose 10k real images and 40k synthetic images as references to calculate the FID.


\subsection{Comparisons with State-of-the-art Methods}\label{sec:sotas}

\noindent\textbf{Baselines.}
We compare our {\thename} with three state-of-the-art methods, 
including EG3D~\cite{Chan2022EG3D}, SDF-StyleGAN~\cite{zheng2022sdfstylegan}, and GET3D~\cite{gao2022get3d}. 
\textit{EG3D} is a 3D-aware image generation method focusing more on the rendered image rather than the geometry.
\textit{SDF-StyleGAN}, which only generates geometry, is a StyleGAN2-based network plus local and global shape discriminators that take as input SDF values and gradients.
\textit{GET3D}, similar to ours, also uses shape and texture branches. 
Its shape branch utilizes a differentiable surface extraction method (i.e. DMTet~\cite{shen2021dmtet})
and its texture branch is based on EG3D.
Different to ours, its supervisions are purely applied on 2D renderings (i.e. images, silhouettes).
The original GET3D~\cite{gao2022get3d} paper is trained in a T-posed RenderPeople dataset. 
We have attempted to train GET3D with our purchased non-T-posed RenderPeople data.
However, the results are poor in terms of geometry and textures,
possibly due to the large pose space and limited training data.
Its results could possibly be improved with more training data, but high-quality 3D data are expensive and difficult to obtain.

For a fair comparison, all these three approaches are trained with our pseudo-GT data. We utilize shape and texture fields to render images to train EG3D and GET3D and get the SDF fields to train SDF-StyleGAN.


\noindent\textbf{Results.}
Tab.~\ref{tab:sota} shows quantitative comparisons and Fig.~\ref{fig:compare_sota} visualizes the results.
Our results beat the others by a substantial margin, especially on COV, MMD, and FID.
We can easily observe the differences from the visual comparisons.

SDF-StyleGAN, which is only able to generate shapes, generates a little better-looking shapes than EG3D. But it often generates disconnected regions so that artifacts are frequently observed in the elbows and ankles. 

EG3D focuses more on rendering high-quality images and cannot produce a 
reasonable geometry sometimes. 
%
The textured mesh does not look good because the geometries are problematic.

%
GET3D results and our results look better since the shapes are clear and contain more details while having less artifacts.
Compared to ours, GET3D generates over-smoothed shapes with more artifacts and its textures are not are sharp as ours. This is also reflected by the FID (54.39 vs 61.69) and FID$_{3D}$ scores (69.70 vs 88.15) in Tab~\ref{tab:sota}. 
Since the geometry generated by EG3D is not accurate enough, the FID$_{3D}$ is much higher (221.73) than other methods.



\begin{figure}[t]
\centering
\includegraphics[width=0.99\linewidth]{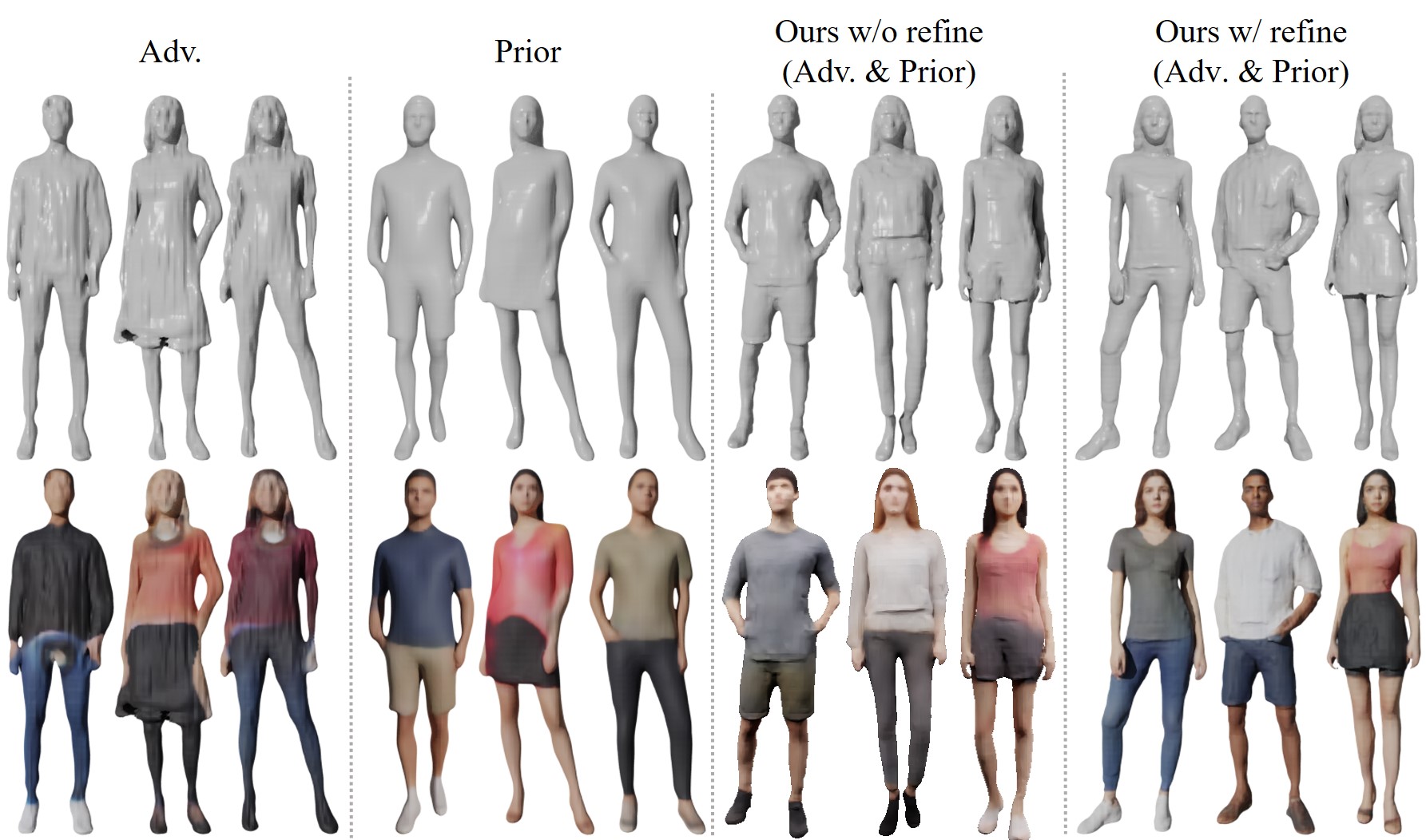}
\caption{
Ablation study on adversarial losses and prior losses.
We compare the human models generated from ``Adv. only'', ``Prior only'', and ''Ours Adv. + Prior''.
(Adv. \& Prior) produces the best results with plausible details on both the shape and the appearance. The refinement module further improves the appearance.
}
\label{fig:gan_prior}
 \vspace{-4mm}
\end{figure}
\begin{table}[t]
\centering
\small
\caption{Quantitative evaluation on adversarial and prior.}
\label{tab:ablation}
{\def\arraystretch{1} \tabcolsep=0.33em 
\begin{tabular}{r|c|c|c|c|c}
\toprule
& COV$\uparrow$ (\%) & MMD$\downarrow$ & FPD$\downarrow$ & FID$\downarrow$ & $\text{FID}_\text{3D}$ $\downarrow$  \\
\midrule
Adv. only &  35.62 & 0.72 & 0.63 & 89.39 &  105.63\\
Prior only &  31.51 & 0.80 & 0.88 & 69.81 &  89.32\\
\midrule
\makecell{Ours (w/o refine)\\(Adv. \& Prior)} &  39.22 & 0.58 & 0.85 &  59.95 & 74.38 \\
\midrule
\makecell{Ours (w/ refine)\\(Adv. \& Prior)} &  \textbf{39.22} & \textbf{0.58} & \textbf{0.85} & \textbf{54.39} & \textbf{69.70} \\
\bottomrule
\end{tabular}
}
\vspace{-4mm}
\end{table}

\subsection{Ablation Studies} \label{sec:Ablation}

\begin{figure}[tb]
\centering	
\includegraphics[width=0.99\linewidth]{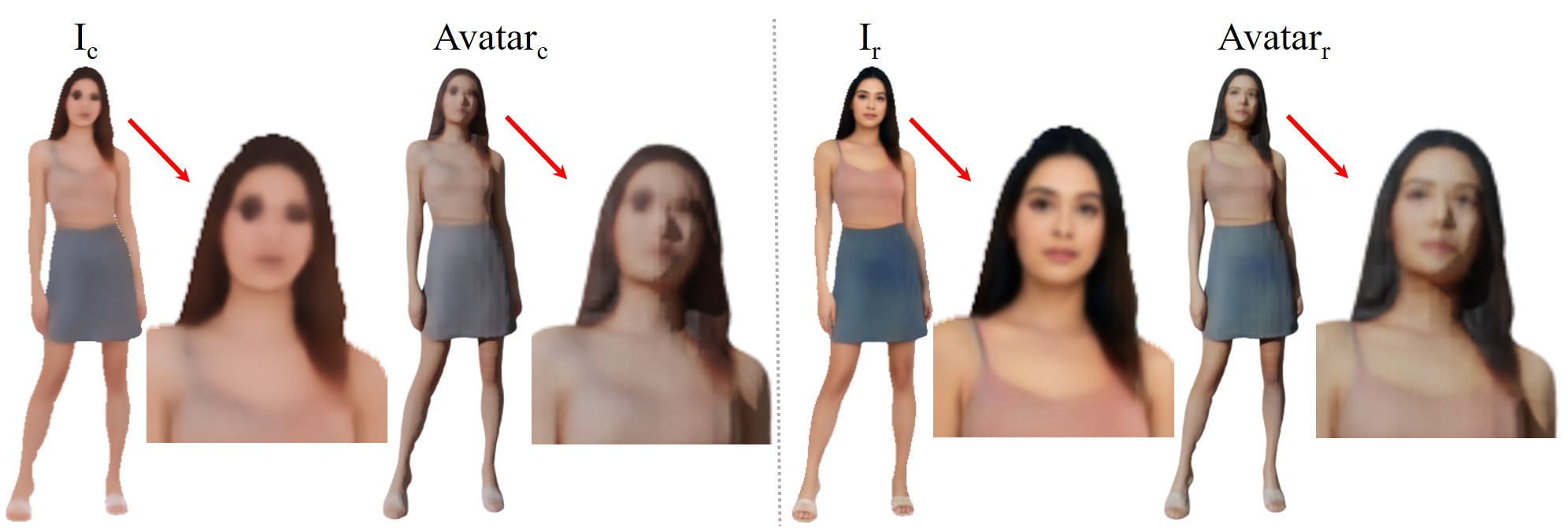}
\vspace{-3mm}
\caption{Ablation on the refine module. 
Using the refinement module brings realistic texture details (right). $I_{r}$ and $I_{c}$ are images directly generated from the network.
Avatars are rendered from texture models by using Blender. 
}
\label{fig:refine}
\vspace{-4mm}
\end{figure}

\noindent\textbf{Ablation study on the adversarial and prior.}
The proposed method consists of two kinds of losses, i.e. adversarial losses and prior losses. 
Each type of loss is able to obtain a certain result.
Ablative experiments, including ``Adv. only'', ``Prior only'' and ``Adv. \& Prior'' with or without refinement, are conducted to demonstrate their effectiveness in Tab.~\ref{tab:ablation}. Fig.~\ref{fig:gan_prior} shows some examples synthesized using random latent code.
Only using adversarial losses in training (Adv. only) easily produces results that contain high-frequency noise, i.e. unrealistic details.
Only using prior losses (Prior only) usually produces overly-smoothed results, which has been observed in 2D image synthesis too.
Ours (Adv. \& Prior) produces the best results with plausible details on both the geometry and the texture. 
When refinement module is included, the final textured mesh further gets a huge improvement in Fig.~\ref{fig:refine}.
%

\begin{figure}[tb]
\centering	\includegraphics[width=0.97\linewidth]{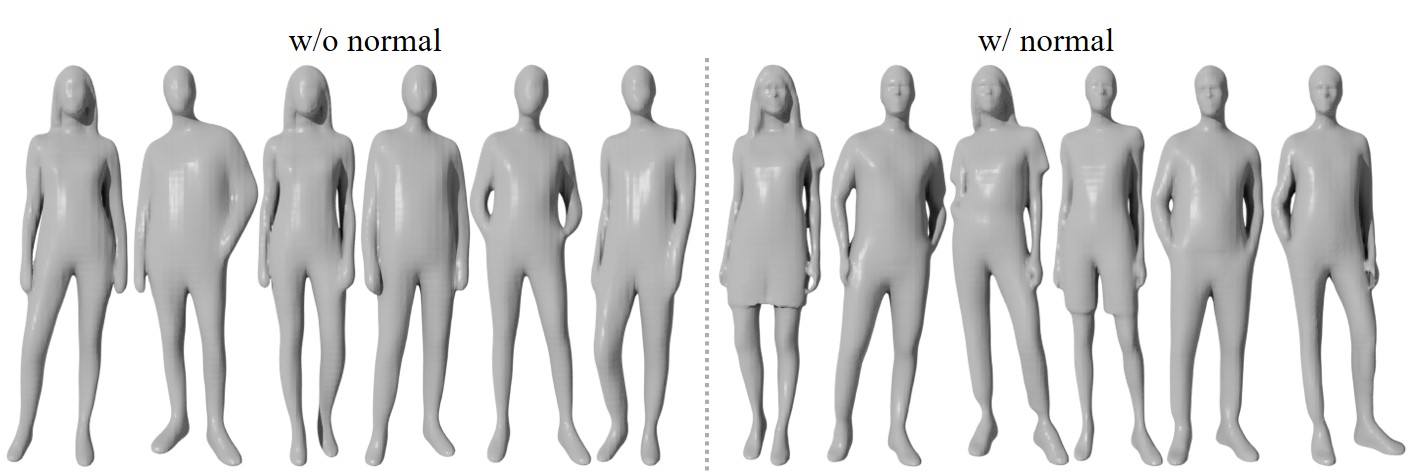}
\vspace{-3mm}
\caption{Ablation study on the intermediate normal supervision. 
Using normal supervision in training brings substantial geometry details (left). 
Note the abrupt depth changes between cloth and skin and the facial geometry.
}
\label{fig:normal_add}
\vspace{-2mm}
\end{figure}

\noindent\textbf{Ablation on normal guidance.}
The explicit use of normal map in 3D reconstruction networks has been demonstrated very helpful~\cite{saito2020pifuhd}.
Thus, we introduce a normal map as guidance for the intermediate feature maps to guide the later geometry generation
and conduct ablative experiments using only prior losses.
Results are shown in Fig.~\ref{fig:normal_add}.
With the help of this normal supervision, more plausible details, especially the separation between upper-/lower-body clothes, emerge on the generated shapes, demonstrating the effectiveness of using the normal map as guidance.

\noindent\textbf{Ablation on refinement module.}
Because the reconstruction prior might be inaccurate and sometimes leads to the unrealistic appearance, especially in the face area. 
To this end, we designed a refinement module. Tab.~\ref{tab:ablation} and Fig.~\ref{fig:refine} illustrate the effectiveness of the refinement module in improving the appearance.

\subsection{Applications}
\label{sec:applications}

\begin{figure}[tb]
\centering	
\includegraphics[width=0.99\linewidth]{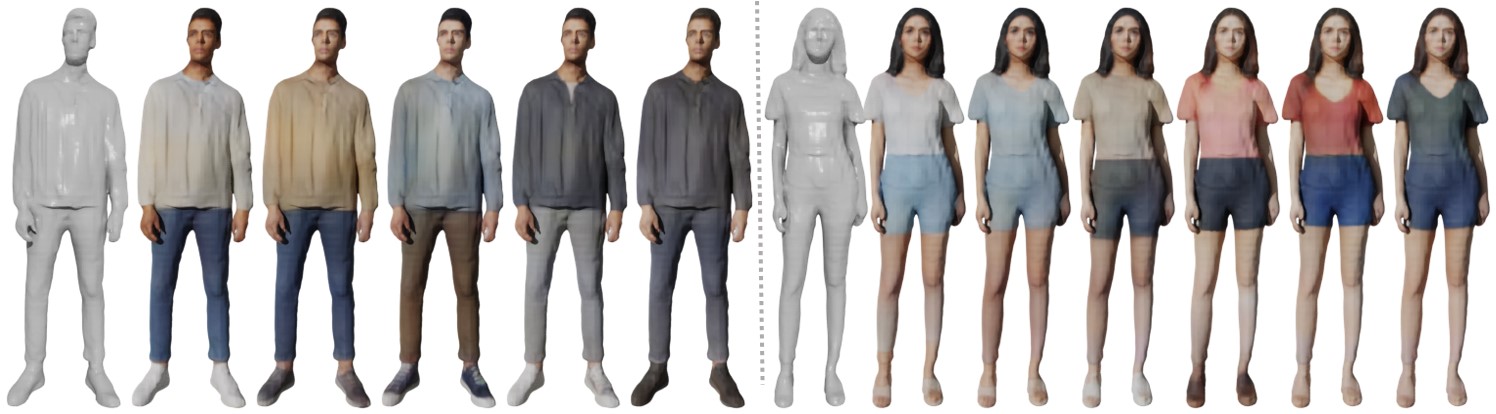}
\vspace{-3mm}
\caption{
Visualization of re-texturing the fixed geometry by different texture latent codes. 
We can see different textures are diverse, plausible, and suitable for the given shape since our texture branch is conditioned on shape branch features.
}
\label{fig:recolor}
\vspace{-4mm}
\end{figure}

\noindent\textbf{Re-texturing the generated meshes.}
The shape/texture latent codes could be different in our two-branch 3D generator.
In fact, we intentionally
include some unpaired shape-texture latent codes (i.e. different values for the shape/texture latent codes) during training and apply adversarial loss on their generations during training.
%
Since the texture branch does not affect the shape branch, the generated shape is completely fixed when the shape latent code is given.
Thus, we can randomly sample texture latent code with the shape latent code fixed, resulting in a useful re-texturing application.
Two examples are shown in Fig.~\ref{fig:recolor}.
Our {\thename} successfully paint the 3D models with different and reasonable texture and takes the conditioned geometry into consideration.

\begin{figure}[tb]
\centering	
\includegraphics[width=0.99\linewidth]{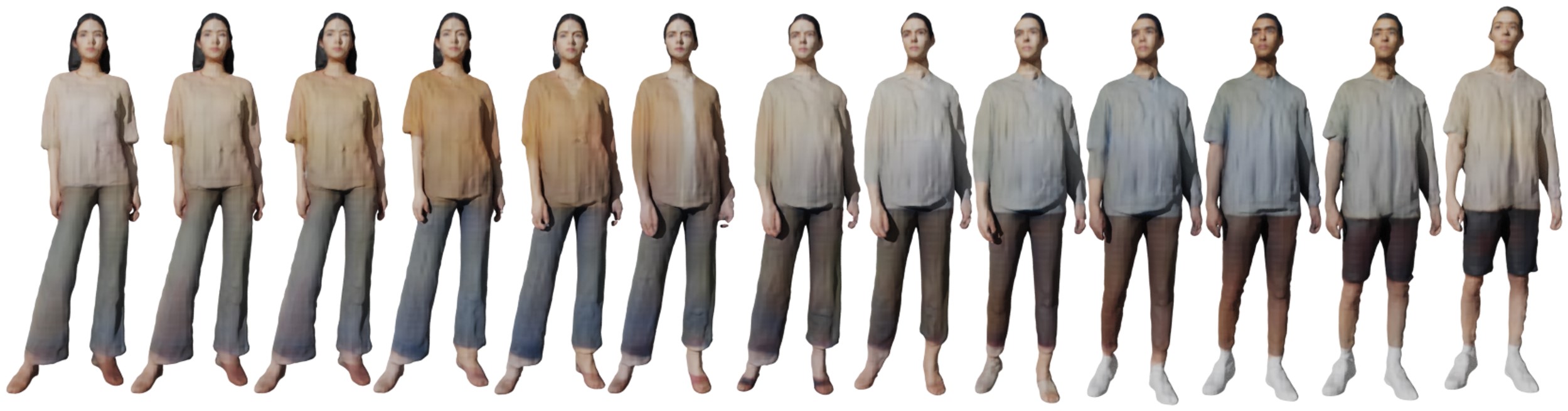}
\vspace{-3mm}
\caption{
Interpolation examples. 
We randomly sample two sets of shape/texture latent codes to generate the right-/left-most examples, then interpolate both the shape and texture latent codes to generate the in-between examples.
}
\label{fig:inter}
\vspace{-4mm}
\end{figure}

\noindent\textbf{Interpolation on shape and appearance.}
With our prior-facilitate learning, our method learns a better structured latent space $z$. 
Thus, we can produce reasonable interpolations given two reference latent codes.
Specifically, we linearly interpolate two random codes and generate textured 3D models accordingly.
Smooth and valid transition models with plausible details are produced as shown in Fig.~\ref{fig:inter}. 

\begin{figure}[tb]
\centering	
\includegraphics[width=0.97\linewidth]{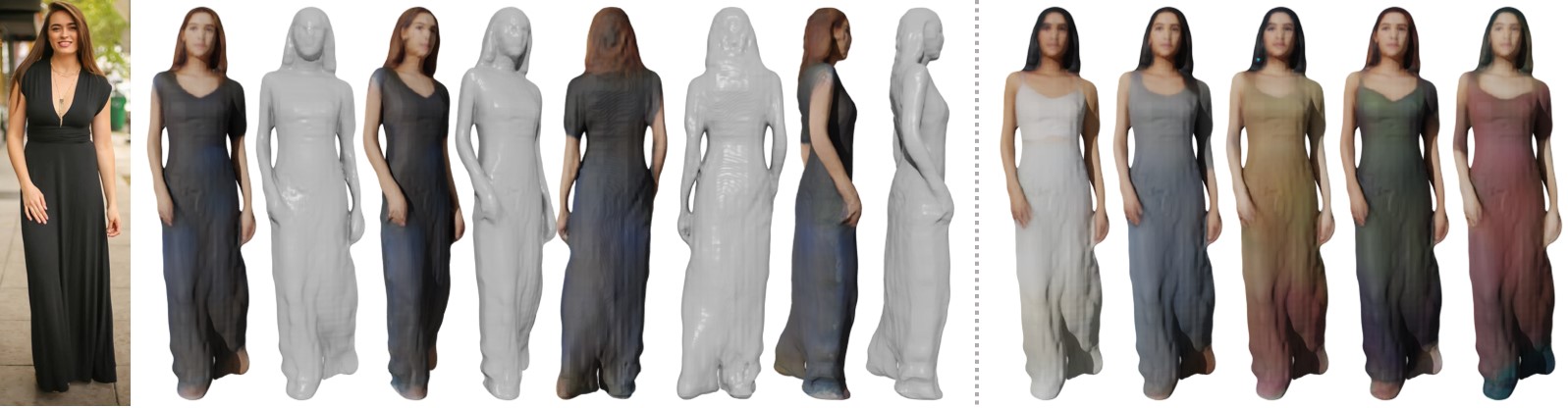}
 \vspace{-3mm}
\caption{
Inversion examples from given reference images.
The reference images are shown on the left, 
inversely optimized human models are shown in different views in the middle,
five re-textured models are shown on the right.
}
\label{fig:inversion}
\vspace{-4mm}
\end{figure}

\noindent\textbf{Inversion results.}
{\thename} is also able to perform image inversion similar to StyleGAN shown in Fig.~\ref{fig:inversion}. 
Given a human image, we first extract its shape/texture field features using the prior network, which are used as the optimization targets to search its corresponding shape/texture latent codes in the latent space.
Note that we use shape/texture field features instead of the reference image when conducting this inversion.
When the optimization is done, a textured human model resembling its reference image is produced.
Compared to previous 3D reconstruction methods (e.g. PIFu), we can easily manipulate these results through latent space editing (e.g. re-texturing, interpolation).

\section{Conclusion} \label{sec:Discussion}
In the paper, We introduced {\thename}, a novel 3D human generator that is able to synthesize diverse and high-quality clothed 3D humans.
It utilizes the priors of the well-trained 2D human generator and 3D reconstructor.  
Numerous experiments have shown that our method greatly outperforms other methods and can support a wide range of applications, including shape interpolation, re-texturing, and single-view reconstruction via latent inversion.

\noindent\textbf{Limitations.}
Our method is only able to synthesize simple standing-pose models, which is restricted by the StyleGAN-Human generator in our prior network.

\clearpage



{\small
\bibliographystyle{ieee_fullname}
\bibliography{egbib}
}

\clearpage

\begin{figure*}
\centering
\includegraphics[width=0.99\linewidth]{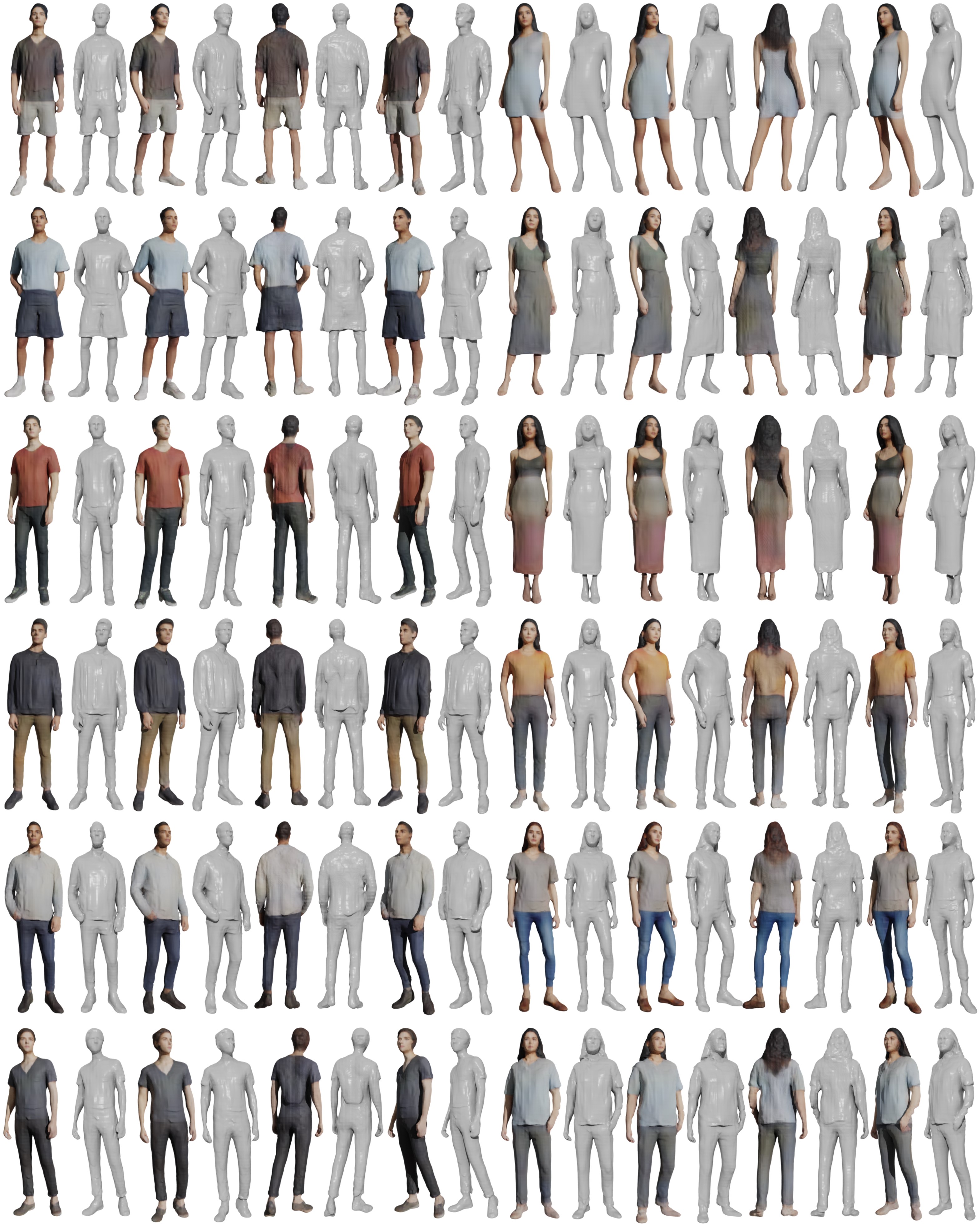}
\caption{
Visualization multi-view images rendering by blender.
}
\label{fig:mul_view}
\end{figure*}

\begin{figure*}
\centering
\includegraphics[width=0.99\linewidth]{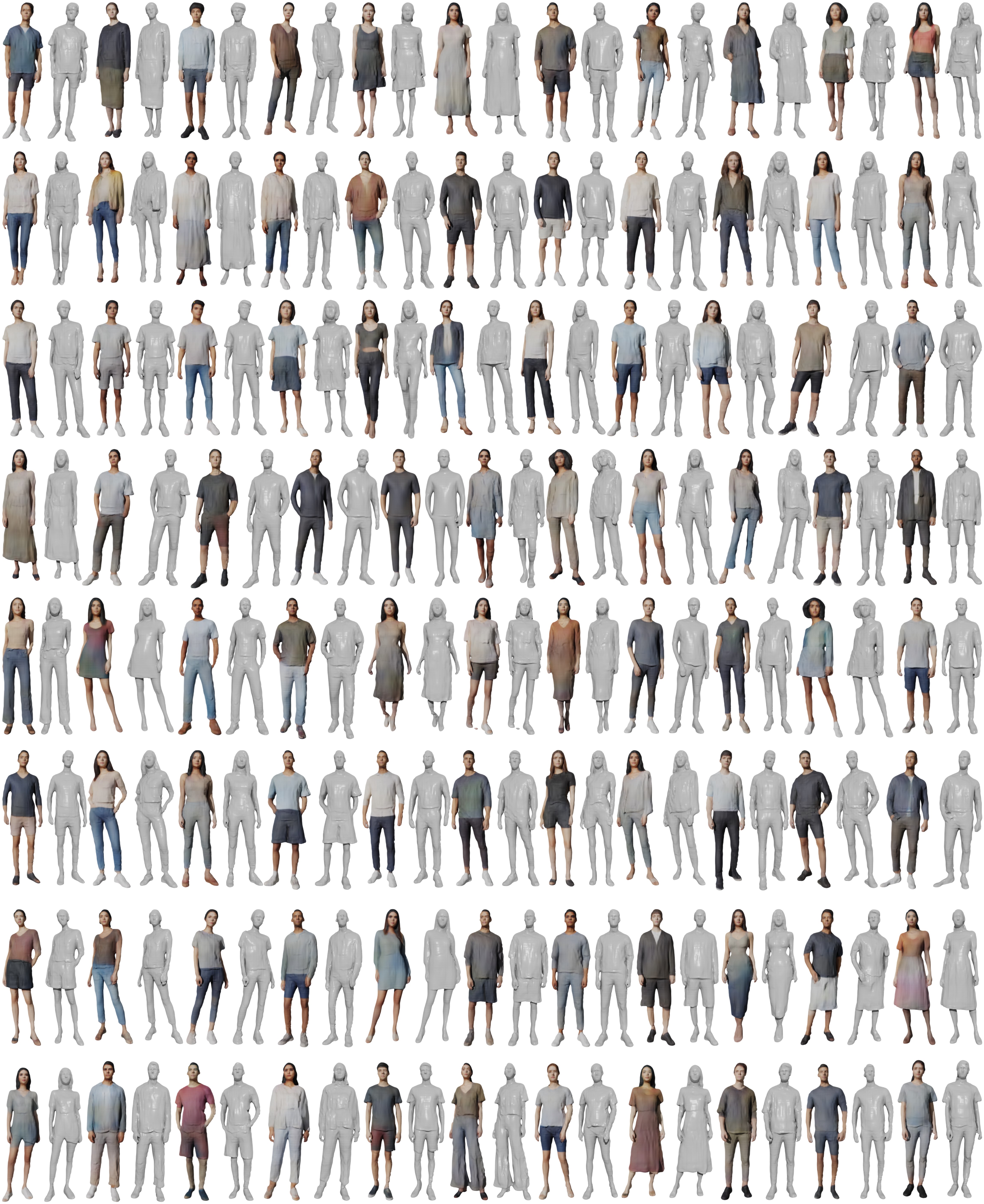}
\caption{
Visualize the results of generated textured mesh.}
\label{fig:sup_sample}
\end{figure*}

\begin{figure*}
\centering
\includegraphics[width=0.99\linewidth]{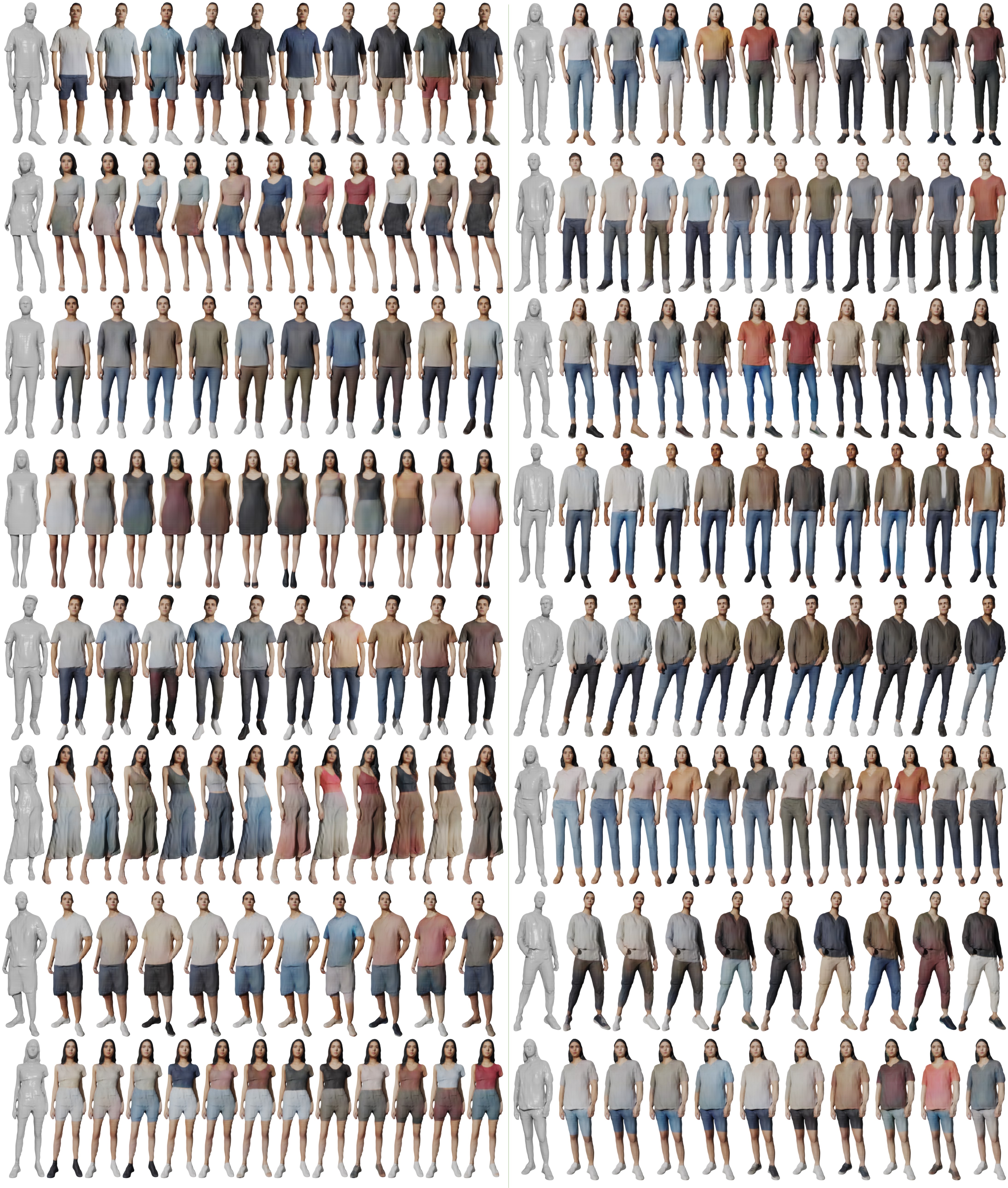}
\caption{
Visualization of re-texturing the fixed geometry by different texture
latent code.}
\label{fig:sup_recolor_all}
\end{figure*}

\begin{figure*}
\centering
\includegraphics[width=0.99\linewidth]{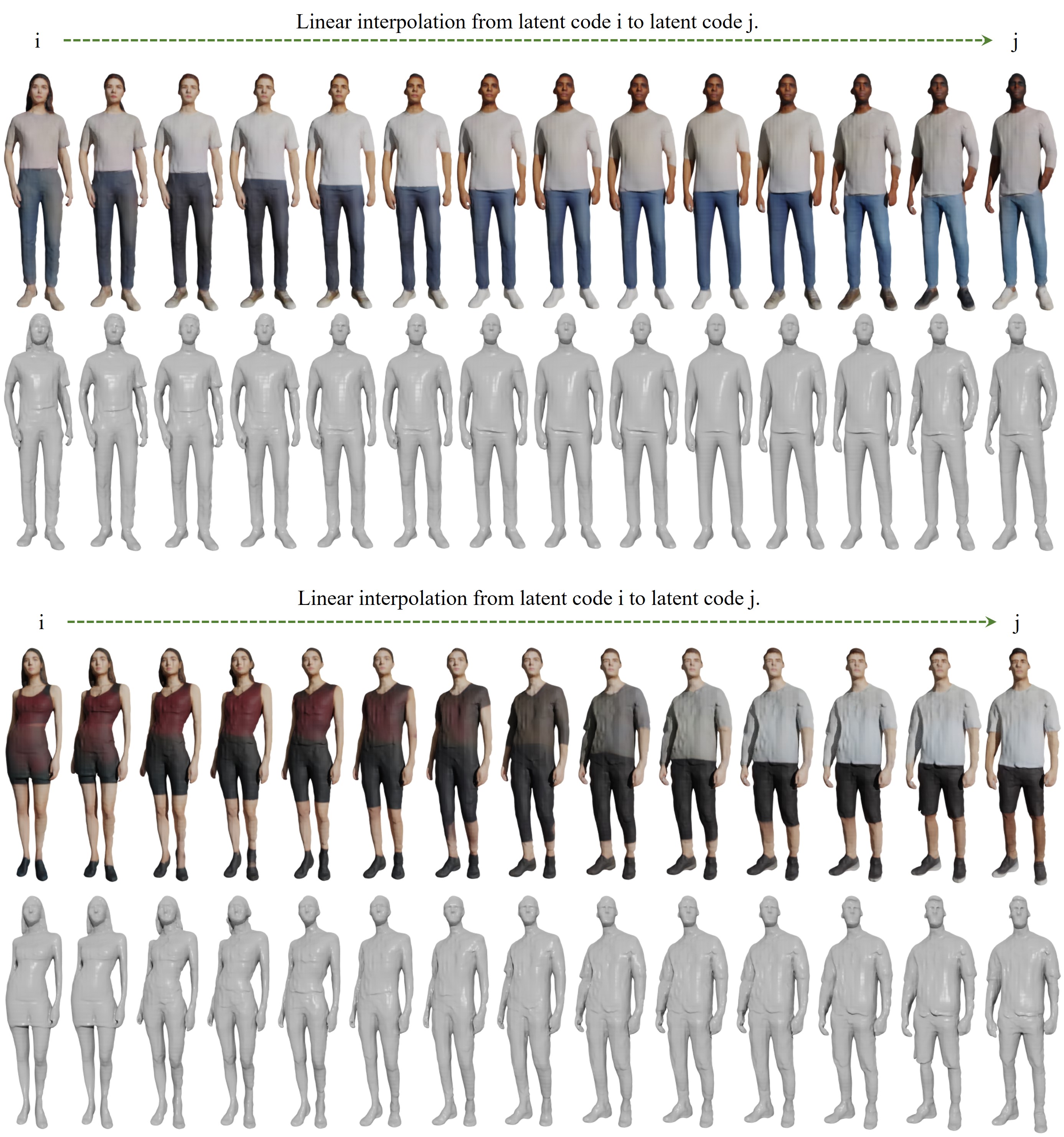}
\caption{
 Interpolation examples. We randomly sample two sets of shape/texture latent codes to generate the right-/left-most examples, then interpolate both the shape and texture latent codes to generate the in-
between examples.}
\label{fig:sup_inter_0}
\end{figure*}

\begin{figure*}
\centering
\includegraphics[width=0.99\linewidth]{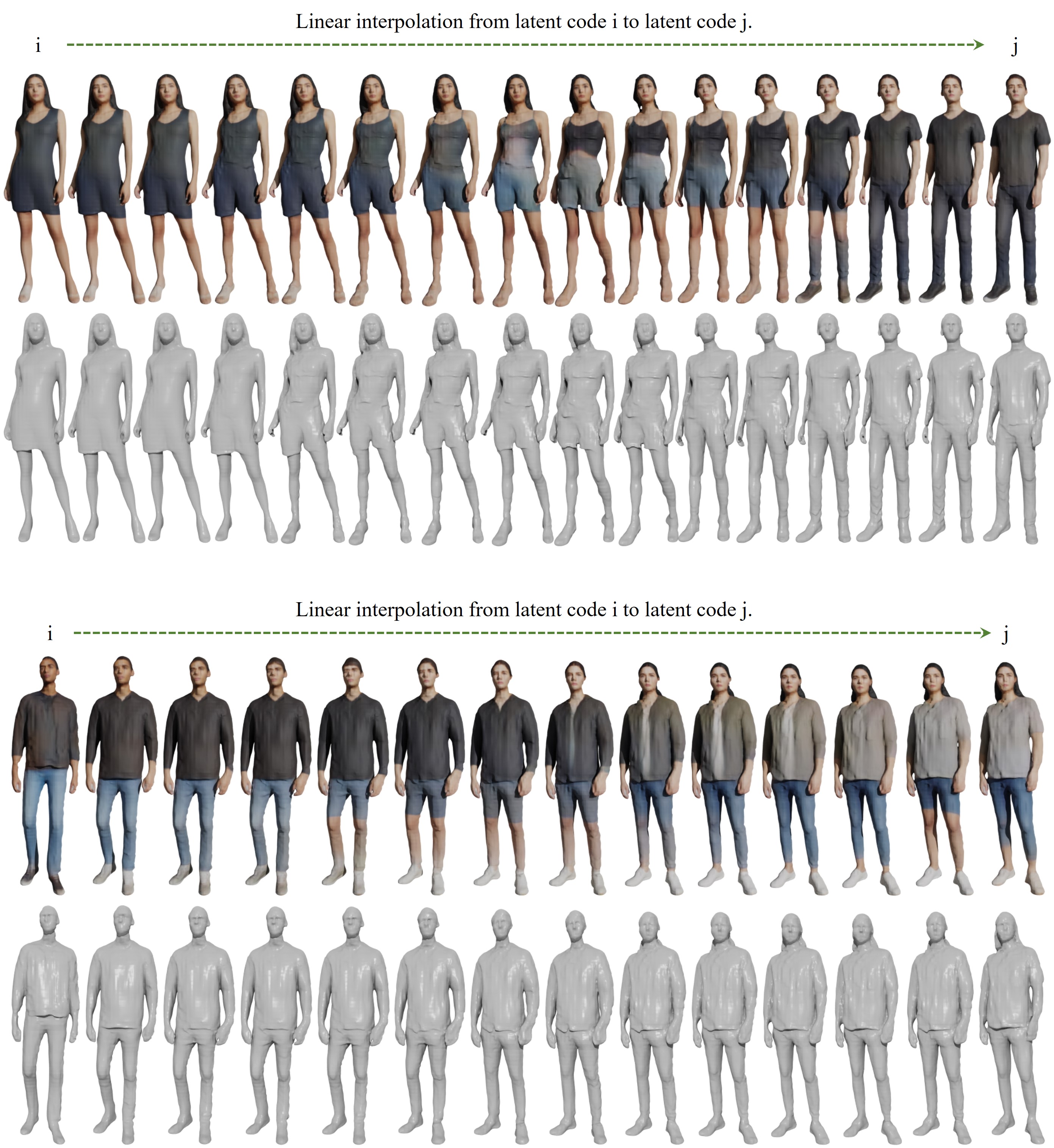}
\caption{
 Interpolation examples. We randomly sample two sets of shape/texture latent codes to generate the right-/left-most examples, then interpolate both the shape and texture latent codes to generate the in-
between examples.}
\label{fig:sup_inter_1}
\end{figure*}

\end{document}